\documentclass[conference]{IEEEtran}
\IEEEoverridecommandlockouts
\usepackage{cite}
\usepackage{amsmath,amssymb,amsfonts}
\usepackage{algorithmic}
\usepackage{graphicx}
\usepackage{textcomp}
\usepackage{xcolor}
\usepackage{subcaption}
\def\BibTeX{{\rm B\kern-.05em{\sc i\kern-.025em b}\kern-.08em
    T\kern-.1667em\lower.7ex\hbox{E}\kern-.125emX}}
\begin{document}

\title{Wait for me! Towards socially assistive \\walk companions  \\
\thanks{
This work has received funding from the European Union's Horizon 2020 framework programme for research and innovation under the Industrial Leadership Agreement (ICT) No. 779942 (CROWDBOT).
}
}

\author{\IEEEauthorblockN{Fernando Garcia}
\IEEEauthorblockA{
\textit{Softbank Robotics Europe}\\
Paris, France \\
ferran.garcia@softbankrobotics.com}
\and
\IEEEauthorblockN{Amit Kumar Pandey}
\IEEEauthorblockA{
\textit{Softbank Robotics Europe}\\
Paris, France \\
akpandey@softbankrobotics.com}
\and
\IEEEauthorblockN{Charles Fattal}
\IEEEauthorblockA{
\textit{CRF La Chataigneraie}\\
Menucourt, France \\
cfattal@lachataigneraie.fr}
}

\maketitle

\begin{abstract}
The aim of the present study involves designing a humanoid robot guide as a walking trainer for elderly and rehabilitation patients. The system is based on the humanoid robot Pepper with a compliance approach that allows to match the motion intention of the user to the robot's pace. This feasibility study is backed up by an experimental evaluation conducted in a rehabilitation centre. We hypothesize that Pepper robot used as an assistive partner, can also benefit elderly users by motivating them to perform physical activity.
\end{abstract}


\section{Introduction}
The effects of the worldwide aging population phenomena have already manifested showing a constant increase in the number of individuals over the age of 65. Specifically, projections are set for 1.5 billion people in 2050 \cite{bremner2010world}. Therefore, it is expected that society will be forced to face a high demand of caregivers and companions for elderly people difficult to fulfill. As a consequence, Socially Assistive Robots are called to help addressing these social needs.

In this case, three main areas of interest for elderly care were identified in \cite{harmo2005needs}. Firstly, securing daily life by warning about dangers, delivering messages and controlling electronic equipment. Secondly, reminding about taking medication, appointments and others. And ultimately, assisting people physically in an unstructured environment. In the last area we frame this study by targeting fundamental research questions -- Could a robotic partner benefit patients in later stages of stroke rehabilitation processes? -- Can elderly individuals be encouraged by a robotic partner to sustain an \textit{active} life-style? -- And which type of operation and working principle should better serve these purposes?

In the present work, Pepper is studied as a care robot for providing guidance and compliant mobility assistance, by adjusting its direction and speed to the user's intention and pace. The sensorless approach presented analyzes the force magnitude and its direction applied to any part of its surface. Also, the system detects obstacles on the way and avoids them by making use of the sensing equipped in the base.

\begin{figure}[h!]
    \centering
    \begin{subfigure}[b]{0.1\textwidth}
        \centering
        \includegraphics[width=\textwidth]{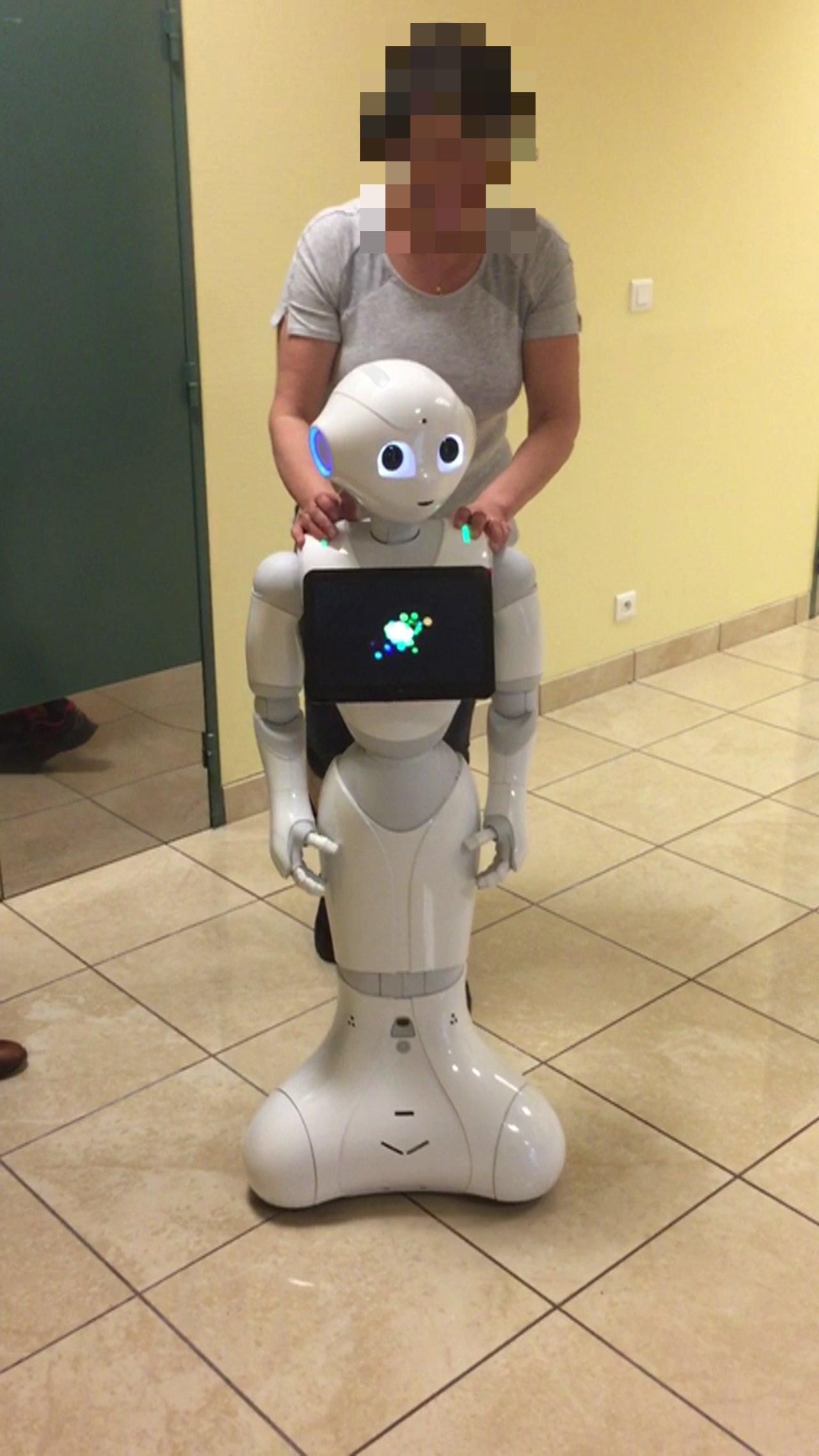}
        \caption{}
    \end{subfigure}%
    ~ 
    \begin{subfigure}[b]{0.1\textwidth}
        \centering
        \includegraphics[width=\textwidth]{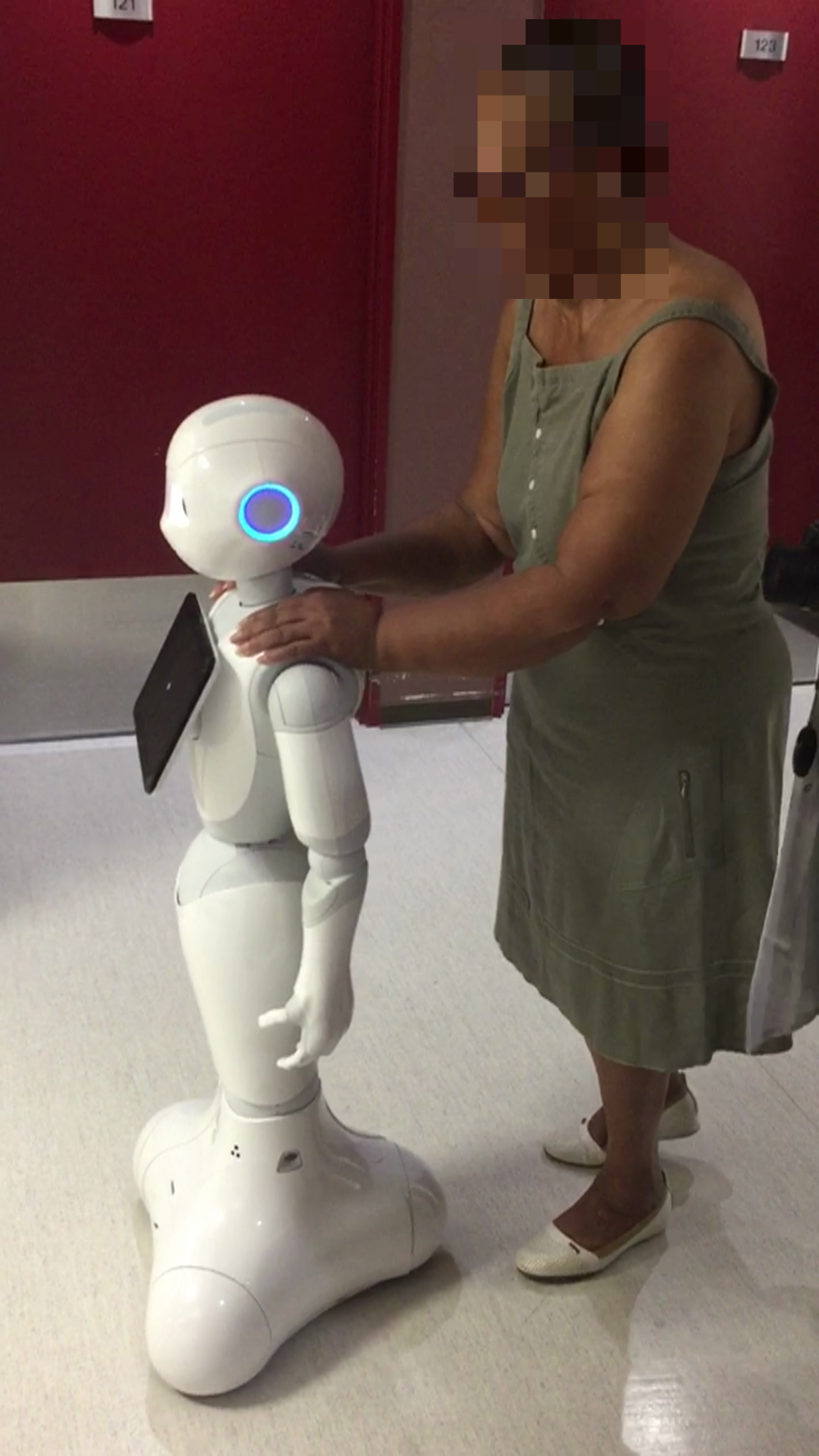}
        \caption{}
    \end{subfigure}%
    ~ 
    \begin{subfigure}[b]{0.1\textwidth}
        \centering
        \includegraphics[width=\textwidth]{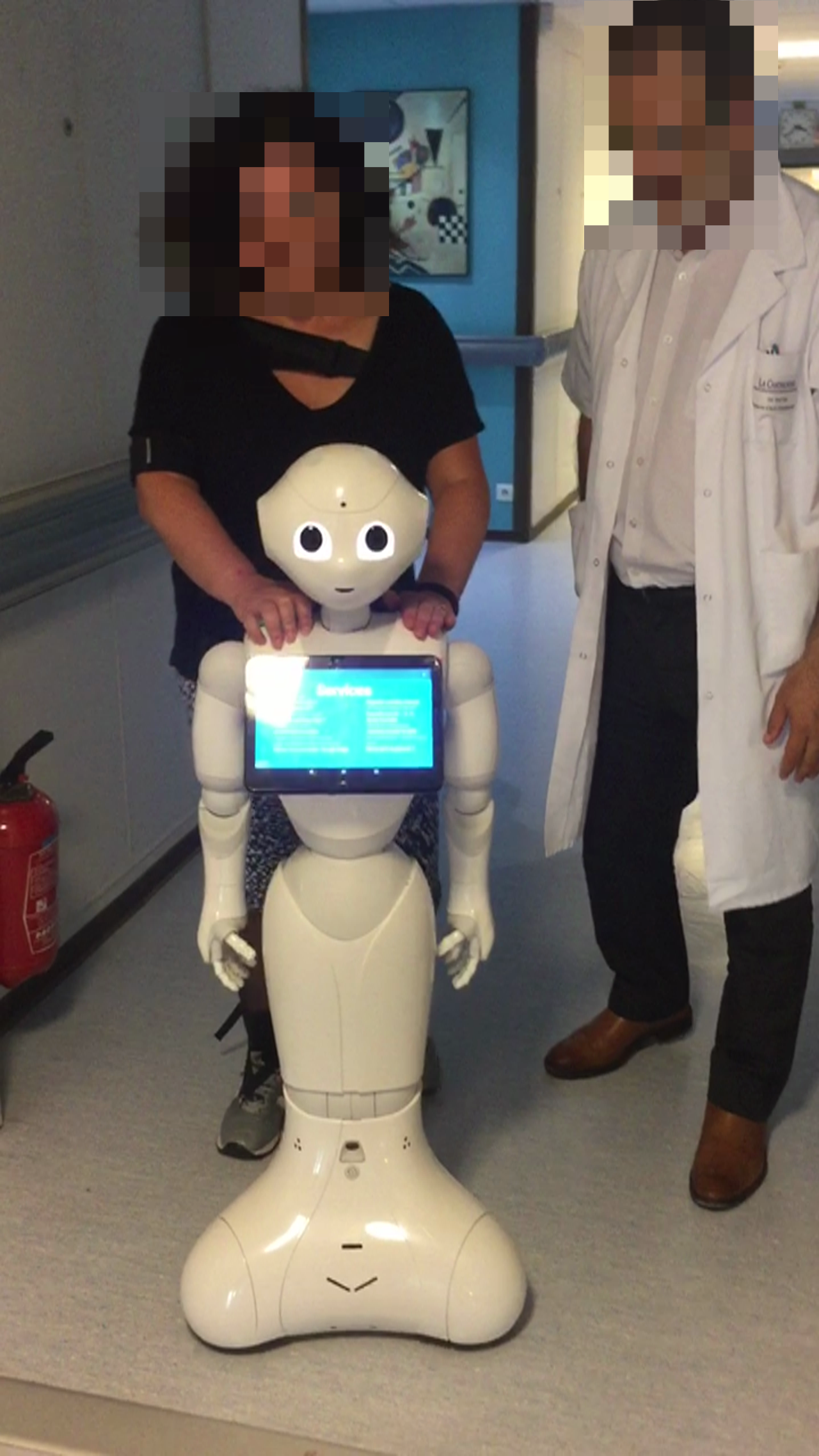}
        \caption{}
    \end{subfigure}%
    ~ 
    \begin{subfigure}[b]{0.1\textwidth}
        \centering
        \includegraphics[width=\textwidth]{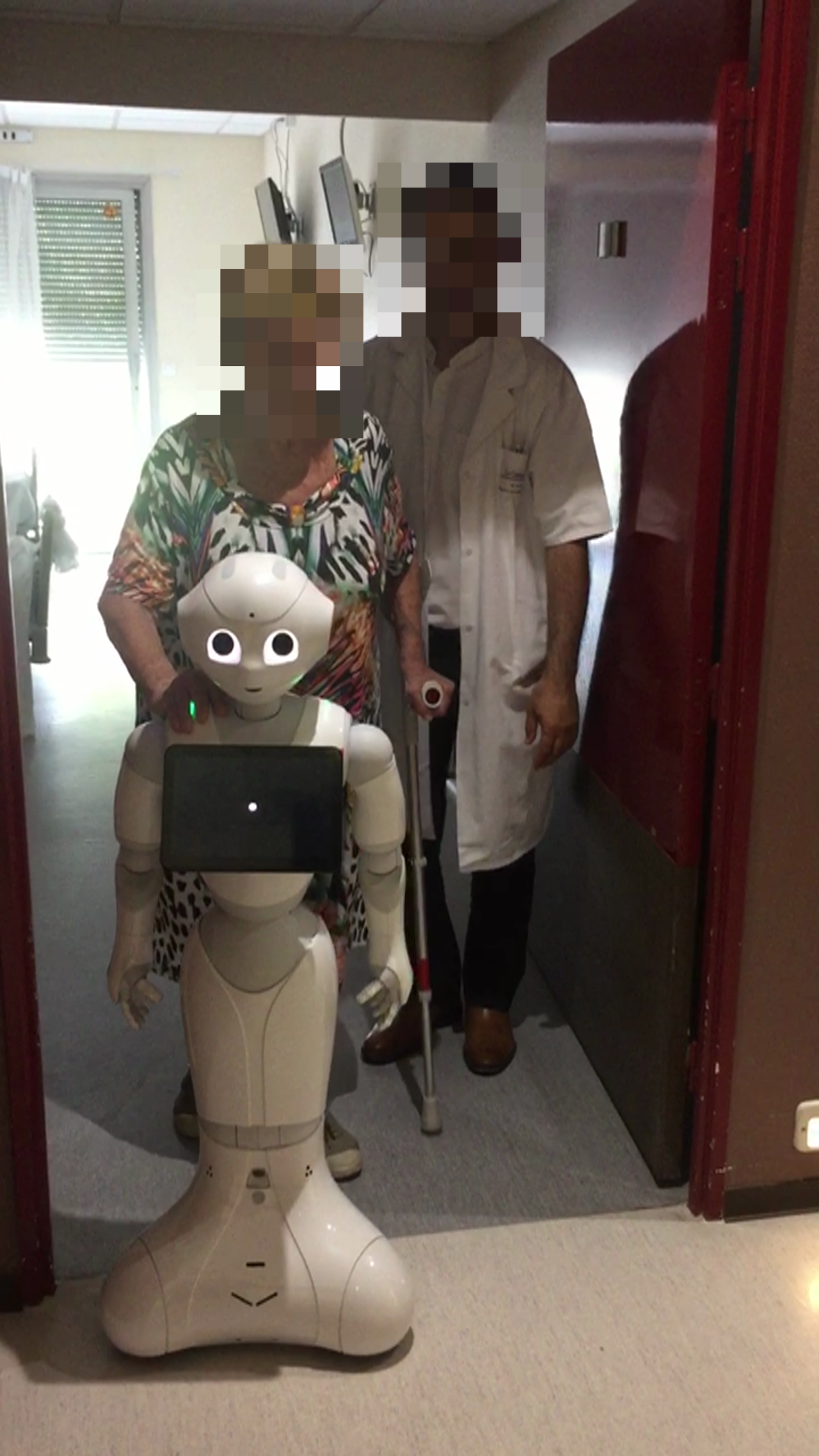}
        \caption{}
    \end{subfigure}
    \caption{Rehabilitation centre experiments carried out by patients with multiple sclerosis (a), parkinson (b), hemiplegia (c) and knee arthroplasty (d) conditions.}
    \label{fig:patients_pics}
\end{figure}

\section{Related work}
Abundant literature has focused on the effects and benefits of social robots in order to reinforce self-confidence \cite{kidd2006sociable}, enhance engagement \cite{winkle2018social}, motivation \cite{deublein2018scaffolding} or promoting healthy habits \cite{schneider2018prec} among elderly. Likewise, assistive technologies are developed to provide solutions for supporting physical activities and rehabilitation therapies: \cite{bedaf2015overview} provides an overview of the proposed systems so far. In fact, researchers are increasingly looking at the personal robots as a solution to extract human motion parameters such as gait, among others, since cameras limit the data acquisition when dealing with different light conditions or clothing.

Feasibility studies which use platforms more socially acceptable, in a physical Human-Robot Interaction (pHRI) context, have inspired this work. Specifically, \cite{piezzo2017feasibility} underlies the key factors that influence a walking trainer robot for elderly individuals, by analyzing the influence of the pHRI with respect to the user's speed. Additionally, \cite{azenkot2016enabling} presents a participatory design approach which enables a PR2 to guide blind people. More recently, \cite{song2017safe} presents a walking-assistant robot that uses gait estimation which is tested with four elderly patients using a semi-humanoid robot. 

However, very few approaches develop physically assistive navigation by the usage of humanoid robots while exploiting its social component. Therefore, further studies need to be conducted considering the needs of chronically ill patients and elderly individuals, by proposing and validating a suitable robotic platform. In this way, investigating the proxemics preferences in this category will allow to provide an effective motivational behaviour, safe model and a system able to capture gait parameters \cite{piezzo2017feasibility}.

\section{Platform suitability assessment}
Pepper is a 1.2 meter tall and 28 kg omnidirectional wheeled humanoid robot platform capable of exhibiting body language, perceiving and interacting with its surroundings, and moving autonomously. In addition, due to its 17 joints and 20 degrees of freedom (DoF) kinematic configuration and edgeless design, the system is suitable for safe HRI \cite{pandey2018mass}. 

The maximum platform speed is 0.8 m/s, which according to the risk evaluation performed using the Head Injury Criterion (HIC) and the New Index for Robots (NIR), establishes a \textit{very low} potential injury level for Pepper robot \cite{cordero2014experimental}. Thus, any risk of a blunt and non-blunt impact is suppressed.

The platform is equipped with a large variety of sensors and actuators that ensure safe navigation. Sensing components include three laser sensors, two sonars and two infrared sensors located in the robot's base, among others. In addition, two tactile detectors on the back of both hands allow human-robot physical awareness, and three bumpers on the base ensure safety in case of sensor failure.

In terms of stabilization, the platform low center of mass mitigates the risk of hard contacts in the upper-body, while increasing its balance. Moreover, when a surface inclination is detected, the torso compensates a possible instability by adapting its position.

The ergonomics of the base shaped on a triangular shape allows the user to locate the feet to both sides of the robot. In addition, the low height of the base with respect to the ground prevents the user from sliding the feet within the gap in between. Finally, Pepper can adapt the position in height by changing the kinematic configuration. Thus, varying joint positions such as knee or hip allows the system to adapt to different heights.


\section{Approach}

An approach for wheel compliance is proposed in this section that consists of three different actuators; front left, front right and back wheels in a holonomic configuration. The sensorless approach for detecting soft pressure in the surface of the robot evaluates the discrepancy between the external displacement and the required to remain in a steady position. Then, direction and intensity responses are transmitted to the wheels resulting in a matching to the user's speed. 

\subsection{User motion intention estimation}
\begin{figure}[h!]
    \begin{center}
        \includegraphics[width=0.45\columnwidth]{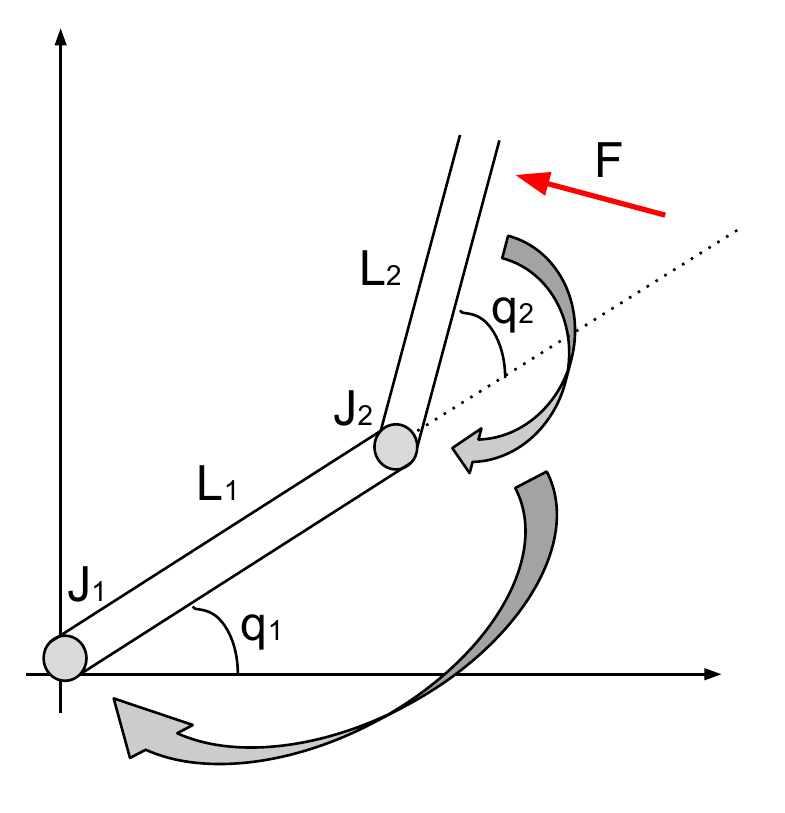}
        \caption{In static conditions, a contact force \(F\) on \(L_2\) is balanced by torques at preceding joints \(i\leq 2\)}
        \label{fig:force_schema}
    \end{center}
\end{figure}

The controller relies on the pressure performed along the robot's surface and transmitted to the closest lower joint. For instance, if an external force \(F\) is applied to the surface of link \(L_2\), the joint \(J_2\) will vary its current state (\(q\)) in terms of current (\(A\)), torque (\(\tau\)), stiffness (\(Nm\)) or others, as can be seen in figure \ref{fig:force_schema}.

Then, in order to maintain the current position and the stability of the system, the joint must generate a torque \(\tau_{total}\) equal to external force \(\tau_{ext}\) and the original joint torque required to keep the desired position \(\tau_{m}\), considering friction \(\tau_{f}\), as shown in equation \ref{eq:tau_ext}:

\begin{equation} \label{eq:tau_ext}
  \tau_{ext} = \tau_{total} - \tau_{f} - \tau_{m}
\end{equation}

Based on equation \ref{eq:tau_ext}, the force applied to the surface \(L_2\) impacts directly the resulting torque \(\tau_{total}\) becoming an estimator of linear force intensity; as per the motion vector, the displacement.
In order to determine the direction of the force to be transmitted to the wheels, an additional component is needed. The angle error \(\theta_{\epsilon}\) in the joint position indicates the direction of the force actuating by computing the discrepancy between the position sensed (s) and the commanded (c), as described in equation \ref{eq:theta_err}:

\begin{equation} \label{eq:theta_err}
  \theta_{\epsilon} = \theta_{s} - \theta_{c}
\end{equation}

Meaning the positive \(\theta_{\epsilon}\), a front force direction, and the negative, a back force direction or vice-versa, depending on the joint. Finally, the vector \(V_m\) shown in \ref{eq:v} can be formalized, and will describe the direction and the magnitude of the motion performed by the wheels:

\begin{equation} \label{eq:v}
  V_m = \begin{pmatrix}\tau_{ext} \\ \theta_{\epsilon} \end{pmatrix}
\end{equation}

\begin{figure}[h!]
  \centering
  \includegraphics[width=0.7\linewidth]{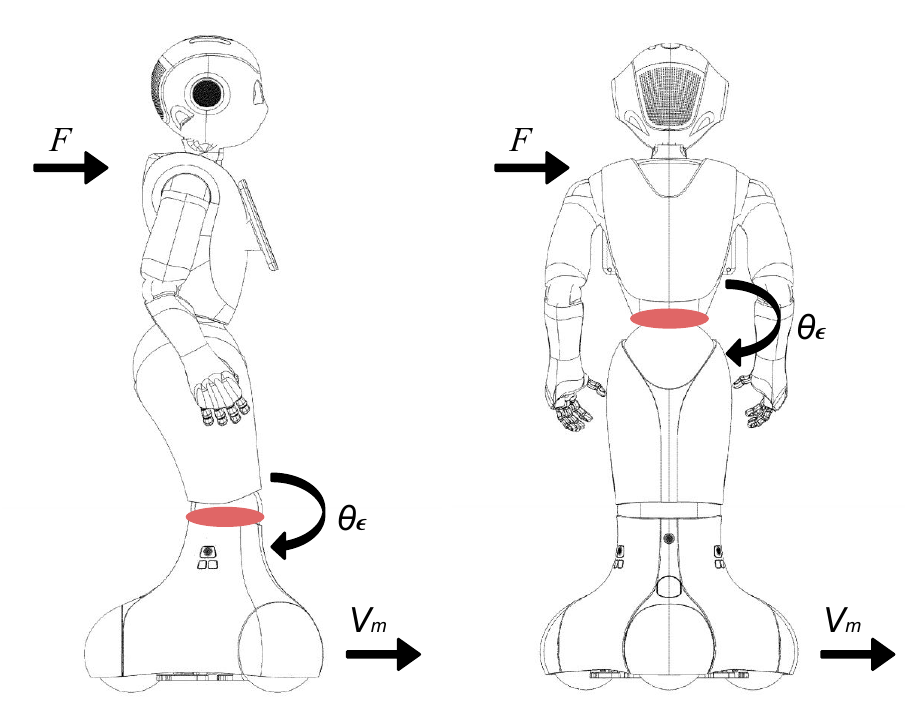}
  \caption{On the left, the displacement of the knee pitch is used for identifying the front/back movements. On the right, the displacement of the hip roll is used for lateral responses.}
  \label{fig:forces_pepper}
\end{figure}

Due to the fact that the robot joints obey an specific rotation motion; roll, pitch and yaw, those need to be used for different directions in order to better estimate the human input induced to the robot. For instance, a displacement in the pitch of the hip can better estimate the command ‘go forward’ rather than the roll of the same joint, since it complies in the natural direction of the force applied, as can be seen in figure \ref{fig:forces_pepper}.

In this case, several joints can be used in order to estimate the \(V_m\) of the robot. In figure \ref{fig:forces_pepper}, the knee pitch is used for forward/backward displacement and the hip roll can be used for rotation over the vertical axis or lateral displacement, as represented.

Then, when a soft pressure is exercised on the body of the robot, assuming the user is located in its back, the reaction is as follows:

\begin{itemize}
    \item North: Translation displacement performed by pushing the robot away from the user.
    \item South: Translation displacement performed by pulling the robot towards the user.
    \item East: Rotation clockwise of the heading, or translation towards the correspondent direction.
    \item West: Rotation counter-clockwise of the heading, or translation towards the correspondent direction.
\end{itemize}

Any combination of the previous motions can be acquired by merging the previous described inputs.
\newline
\subsection{Degree of compliance}
The system can be operated with the obstacle avoidance sensors located in the base enabled so that the displacement is always safe and any risk of collision gets mitigated. If desired, based on the previous information about the location of the surrounding obstacles, the system degree of compliance can be modified during the execution, establishing a logistic relation with respect to the distance of a known obstacle. 

Consider the robot \(R_t(x,y)\) moving at a constant speed at time \(t\) towards an static obstacle \(O_t(x+2,y)\) with euclidean distance \(d_t(R,O)=2\), being \(R\) and \(O\) located in the world coordinate system. At time \(t+1\), the system will be at \(R_{t+1}(x+1,y)\) if and only if \(d_{t+1}(R,O)\geq1\), allowing a maximum degree of robot compliance \(R_c\). However, at time \(t+2\), the static obstacle \(O_{t+2}(x+2,y)\) in the field of view of the robot at \(d_{t+2}(R,O)<1\) will exponentially decrease \(R_c\) in the robot's vector of motion \(V_m(1,y)\).

The relation between \(d(R,O)\) and \(R_c\) is formalized as a generalized exponential function, see equation \ref{eq:f_x} which can be simplified as equation \ref{eq:R_c}.

\begin{equation} \label{eq:f_x}
  f(x) = A + \cfrac{K-A}{(C+e^{-B(x-M)})^{1/v}}
\end{equation}

where, \\

\noindent
\textit{A} = Lower compliance limit, 0 \\
\textit{K} = Upper compliance limit, 1 \\
\textit{x} = Distance between robot and obstacle d(R,O) \\
\textit{v} = Frequency growth in the compliance limits, 1 \\
\textit{M} = Set middle point of the function, 1 \\
\textit{B} = Growth rate \\

becoming,

\begin{equation} \label{eq:R_c}
R_c(d) = \cfrac{1}{1+e^{-B(d-1)}}
\end{equation}

\subsection{Control loops}
\begin{figure}[h!]
    \centering
    \begin{subfigure}[b]{0.17\textwidth}
        \centering
        \includegraphics[width=\textwidth]{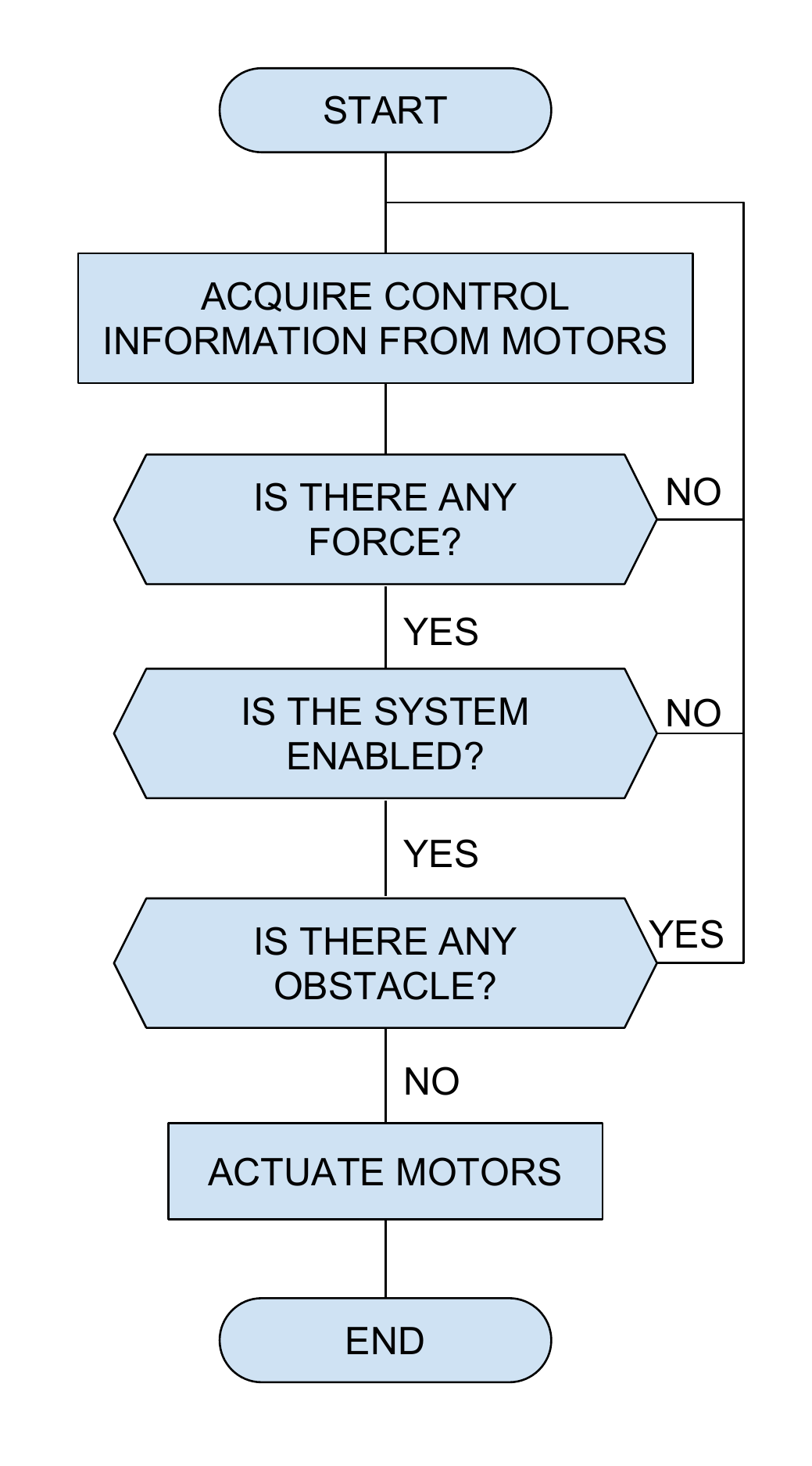}
        \caption{}
    \end{subfigure}%
    ~ 
    \begin{subfigure}[b]{0.17\textwidth}
        \centering
        \includegraphics[width=\textwidth]{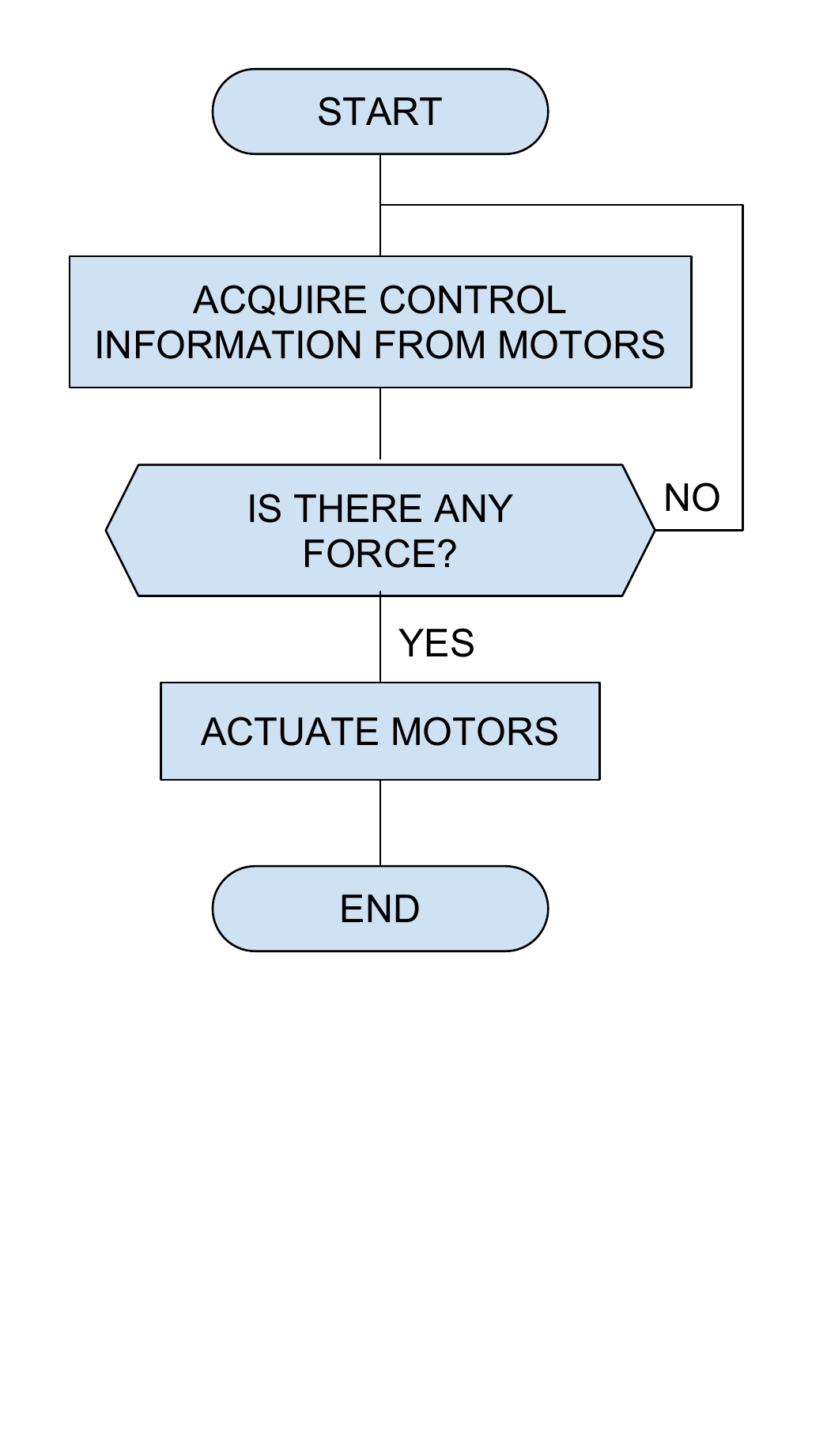}
        \caption{}
    \end{subfigure}
    \caption{Robot control loop. (a) Assisted (b) Non-assisted}
    \label{fig:control_loop}
\end{figure}

The system is equipped with an additional operation device that let lock and unlock the motion to a superuser by the use of touch sensors. It can operate in two different control loops: assisted and non-assisted (see figure \ref{fig:control_loop}). In the first case, the required information for torque computation is retrieved so that if $\tau_{ext} > K$, the system is enabled by a superuser and there are no obstacles detected, the correspondent command is sent to the actuators. In the second case, the robot complies with the user intention without considering any other input.

\section{Experimental evaluation}
In order to validate the initial stage of the implementation and collect user feedback, several trials were performed by four patients with different medical diagnostics; multiple sclerosis, parkinson, hemiplegia and knee arthroplasty were the user conditions. The experiments were carried out under constant supervision of a medical team in a rehabilitation centre. 

Little instructions were provided to the participants; only the working principle of \textit{pushing/pulling} and the back of the robot as suggested location for operating it, as can be seen in figure \ref{fig:patients_pics}. Patients performed an activity 30 minutes long, in average, pushing the robot's back to transmit their movement intention at their own free will. 

Patients displaced the robot on a straight line in corridors, but also laterally in few corners using two side-motion approaches. The first one allowed side translations with a fixed heading. The second one replaced the translation for a rotation over the vertical axis of the humanoid.

Due to the sensitivity of the users involved in the experiment, the approach presented was run standalone with a speed limit of 0.35 m/s and a maximum acceleration of 0.3 m/s$^2$. Furthermore, the robot's kinematic configuration was locked to remain standing in order to minimize any risk during execution.

\section{Preliminary results}

Here we present the feedback and observations collected during and after the user experiments from both patients and medical team. In simple cases such as robot's maximum speed, motion constraints or pressure sensitivity, suggestions were incorporated \textit{on the fly} in order to maximize the user experience and deepen the feedback.

Overall, after a training period not higher than 5 minutes, the four patients were able to operate the robot successfully and all of them manifested a positive feedback towards the concept. After that and while moving with ease, the feedback was constantly provided.

In terms of motion flow, patients expressed their preference for the robot to rotate over its vertical axis instead of a lateral translation. In addition, the medical team suggested to disable the back displacement for two reasons; mitigating any undesired risk of contact with the patient, and better supporting the stabilization of the patient by remaining still. Moreover, in contrary to our hypothesis the triangular base was not perceived as the \textit{feet friendly} approach: nearly all of the patients asked for the possibility to reverse the orientation of the base in order to push the front shoulders.

According to our observations, patient \ref{fig:patients_pics}(c) found difficult to lean on the robot, suggesting a need for posture personalization. Moreover, the condition exhibited did not allow the user to perform a fluent pressure on the robot's shoulders preventing a smooth displacement. However, the user was able to find an alternative way to transmit its intention to the robot by applying the same pressure on the neck joint instead of the shoulders. In case of patient \ref{fig:patients_pics}(d), the use of a crutch and the robot at the same time, revealed a need for an autonomous mode where the robot is able to navigate close to the user with no need for physical contact.

At the technical side, few issues were identified; the most relevant one, related with the overheating of the robot's knee joint. In some cases, that prevented the user to perform the activity uninterruptedly and the system needed to be restarted. The reason behind this phenomena was identified as the substantial amount of weight some participants overloaded the platform during execution. In addition, this uncovered the inter-user variability in terms of pressure which different users apply to the system.

\section{Conclusions \& Future work}
In summary, given the inter-user variability of force applied over the surface of the robot, the personalization of the system response based on the continuous input of the user would allow to account, not only for the particularities of each interaction but also for the different physical constraints of each patient.

Moreover, additional trials at this stage are necessary in order to ensure the robustness of the gate estimation approach by providing a quantitative analysis of the user's pressure to better hypothesize motion intention. Then, in later stages of the development process, a shared control navigation approach that aims to find merging strategies for the user intention and the robot will, based on a predefined set of physical objectives defined by a medical team, will be implemented.

Despite the encouraging preliminary results in terms of feasibility and positive user response, the evaluation of the benefits of the system for motivating light-physical activity in elderly individuals, and patients with reduced mobility during later stages of a rehabilitation process, remains as a subject of deeper study. In this regard, we plan to conduct a long-term user evaluation at the end of the implementation phase. 

\section*{Acknowledgment}
We would like to thank the staff of the CRF La Chataigneraie Menucourt for its collaboration during the experimentation.

\bibliographystyle{splncs04}
\bibliography{HRI2019}
\end{document}